# Autonomous labeling of surgical resection margins using a foundation model

## Authors

Xilin Yang[1,2,3], Musa Aydin[1,4], Yuhong Lu[1], Sahan Yoruc Selcuk[1,2,3], Bijie Bai[1,2,3], Yijie Zhang[1,2,3], Andrew Birkeland[5], Katjana Ehrlich[6], Julien Bec[6], Laura Marcu[6,7], Nir Pillar[1,2,3,8*], Aydogan Ozcan[1,2,3,9*]

## Affiliations

[1]Electrical and Computer Engineering Department, University of California, Los Angeles, CA, 90095, USA.

[2]Bioengineering Department, University of California, Los Angeles, CA, 90095, USA.

[3]California NanoSystems Institute (CNSI), University of California, Los Angeles, CA, 90095, USA.

[4]Department of Computer Engineering, Fatih Sultan Mehmet Vakif University, Istanbul, Turkiye

[5]Department of Otolaryngology—Head and Neck Surgery, University of California, Davis, CA, 95817, USA

[6]Department of Biomedical Engineering, University of California, Davis, CA, 95616, USA

[7]Department of Neurological Surgery, University of California, Davis, California 95616, USA

[8]Department of Pathology, Hadassah Hebrew University Medical Center, Jerusalem, Israel

[9]Department of Surgery, University of California, Los Angeles, CA, 90095, USA.

*nir.pillar@mail.huji.ac.il

*ozcan@ucla.edu

## Abstract

Assessing resection margins is central to pathological specimen evaluation and has profound implications for patient outcomes. Current practice employs physical inking, which is applied variably, and cautery artifacts can obscure the true margin on histological sections. We present a virtual inking network (VIN) that autonomously localizes the surgical cut surface on whole-slide images, reducing reliance on inks and standardizing margin-focused review. VIN uses a frozen foundation model as the feature extractor and a compact two-layer multilayer perceptron trained for patch-level classification of cautery-consistent features. The dataset comprised 120 hematoxylin and eosin (H&E) stained slides from 12 human tonsil tissue blocks, resulting in ~2 TB of uncompressed raw image data, where a board-certified pathologist provided boundary annotations. In blind testing with 20 slides from previously unseen blocks, VIN produced coherent margin overlays that qualitatively aligned with expert annotations across serial sections. Quantitatively, region-level accuracy was ~73.3% across the test set, with errors largely confined to limited areas that did not disrupt continuity of the whole-slide margin map. These results indicate that VIN captures cautery-related histomorphology and can provide a reproducible, ink-free margin delineation suitable for integration into routine digital pathology workflows and for downstream measurement of margin distances.

## Introduction

In surgical oncology, achieving complete removal of the tumor at the primary site is the desired operative goal, and removing all microscopic or subclinical foci of malignancy is essential and critical to achieving successful local control of the disease[1,2]. During surgery, the tumor is resected along with a narrow strip of normal tissue to ensure complete removal of the tumor[3]. Most commonly, tumor resection is performed using electrocautery, which applies high-frequency alternating current to generate heat in tissues for surgical excisions or to provide hemostasis to any severed blood vessels. Hence, the cauterized tissue delineates the resection margin and separates the resected tissue from the patient. A surgical pathologist examines the entire specimen grossly and microscopically to determine whether the surgical margins are free or involved with residual tumor. If the surgical margins are free of residual tumor, without evidence of significant epithelial dysplasia, the pathological report will assign the resected tumor as completely excised. However, if cancer cells are identified at the resection margin, the tumor is considered incompletely excised, indicating the presence of residual mass *in situ*. It has been shown that failure to achieve a clear surgical margin results in an increased risk of local recurrence and a subsequently reduced chance of survival[4,5]. Adjuvant treatment (i.e., requirement of chemotherapy or radiotherapy or both) may be advised based on the pathology report, emphasizing why accurate margin assessment is crucial for appropriate patient care[6].

Despite its critical importance, accurate margin assessment is plagued by several procedural limitations[7]. First, tissue processing and slide preparation introduce artifacts and variability. For instance, the inking of margins, a manual procedure done by histotechnicians, is human-dependent and prone to high procedural variability. Pitfalls include ink penetration into deeper tissues, ink washout, and the marking of non-margin aspects of the specimen. Furthermore, after formalin fixation, tissue shrinkage and distortion can obscure margin identification, and some resection margins cannot be accurately labeled. A second significant challenge is that cautery-related thermal injury at the cut surface can blur glandular/epithelial contours and produce nuclear streaming and smudging that mimic dysplasia or residual carcinoma, complicating edge interpretation and contributing to discordant margin calls[8]. Finally, there are cases where no margin tagging is performed at all, such as in small resections where there is no reliable way to macroscopically identify the margins, or in lesions considered by the surgeon to have low malignant potential. In such cases without inking, microscopic examination can unexpectedly reveal areas with carcinoma or high-grade dysplasia, complicating the subsequent follow-up care plan.

Motivated by these drawbacks, here we demonstrate an alternative, reliable method for histologic margin assessment. We present an autonomous labeling algorithm, dubbed Virtual Inking Network (VIN), which utilizes a foundation model-based method to digitally mark tissue resection margins. We first utilized a frozen foundation model, the Hierarchical Image Pyramid Transformer (HIPT)[9], as a feature extractor to get sparse embeddings of each histology image patch. Then we optimized a compact two-layer Multilayer Perceptron (MLP)[10] to classify the extracted features into whether that patch contains cauterized boundaries or not. We used a total of 120 whole-slide images (WSI), with 100 used for training and 20 for blind testing, totaling ~2 TB of uncompressed raw image data. The testing WSIs were sourced from different tissue blocks that the model had never seen in the training process to ensure generalization. VIN successfully predicted cauterization boundaries on unseen images, in concordance with a board-certified pathologist. Operationally, VIN's automated margin map expedites slide orientation and enables targeted inspection at the tissue edges. It also furnishes consistent measurements of margin length and proximity, even when specimen inks or sutures are missing. Because cautery changes are a common visual signature across resections, the HIPT-derived representations used by VIN are well-suited to scale and can be integrated upstream of other computational tasks (e.g., dysplasia or tumor detection) to constrain search regions and reduce false positives at tissue edges, thus complementing existing pathology workflows that determine whether surgical margins are free or involved with residual tumor and guide adjuvant treatment decisions.

## Results

Figure 1 outlines our automated approach and the traditional physical margin labeling in surgical practice. In routine clinical workflow, tissue is excised using electrosurgery; immediately after specimen removal (Fig. 1a), the resection surface is typically marked with colored inks at the grossing bench to designate the true surgical margins. After formalin fixation, tissue embedding, processing, and sectioning, the ink carries through to H&E sections, where it delineates the margin under brightfield microscopy (Fig. 1b). However, the inking process has practical drawbacks: inks may be variably applied or incompletely retained; interpretation at the tissue edge is challenging in small resections where orientation is easily lost, and tagging is often omitted; and cautery-related thermal injury can obscure epithelial and stromal architecture at the cut surface. To mitigate the limitations of the manual inking process, we implemented a digital, autonomous labeling workflow (Fig. 1c) in which the VIN produces a virtual delineation of cauterization boundaries directly on the WSI, eliminating dependence on physically applied labels and standardizing margin-focused review. VIN uses a frozen foundation model-based encoder as the feature extractor and a compact two-layer MLP as the decision head. We generated sparse embeddings for individual histology patches and trained the classifier to identify patches containing cautery-consistent boundary features. For model development, we assembled 120 H&E-stained WSIs from 12 tissue blocks with experimentally induced *in situ* cauterization, digitized all the resulting H&E slides, and obtained board-certified pathologist annotations of the cut-surface boundary as ground truth (see the Methods). We formulated the task as patch-level binary classification, using high-resolution image patches as input and a single label indicating the presence or absence of cautery boundary features as output. The model was trained on 100 WSIs, with 20 WSIs held out for blind testing; to ensure generalization, the test slides were drawn from two completely new tissue blocks not used during training. During inference, patch-level predictions are aggregated into a continuous boundary prediction with voting across the WSI and rendered as a virtual ink overlay, facilitating rapid visual confirmation and precise edge-focused review by a pathologist (Fig. 1c; see Methods section for data preprocessing and visualization details).

In Figure 2, we present a blind evaluation of VIN on two unseen WSIs from a new tissue block. The left panels show the input H&E brightfield images; the middle panels display VIN's predictions, with cauterized cut-surface regions rendered in red and non-cauterized tissue in blue; the right panels overlay the pathologist-provided ground truth on the corresponding inputs. For consistency with the expert review, we use an additional green label to denote areas the expert judged equivocal. These equivocal regions were excluded by design from both training and testing to avoid label ambiguity and to ensure that performance estimates reflect only definitive edge calls. Overall, predicted boundaries align closely with the expert annotations, showing good agreement along the cautery-affected interface for the entire WSI. Two slides from different levels of the same tissue block are shown in Figure 2a and Figure 2b, illustrating VIN's robust performance on independent tissue sections.

To further assess generalization, we conducted an additional blind test on two additional WSIs from a different tissue block (also excluded from training) and visualized the results in Figure 3 using the same layout as in Figure 2. In this slide set, we observed a false-positive boundary in the left mid-portion of the section, with a small adjoining region demonstrating true-positive predictions; nonetheless, the overall correspondence to expert labels remained high across most of the margin. Beyond Figures 2 and 3, six additional slides, three from each of the two tissue blocks, are compiled in Supplementary Figure 1. In these additional sections, VIN's outputs are consistent across serial cuts from the same block, despite the expected cellular-level heterogeneity, supporting the notion that the model is leveraging cautery-related histomorphologic signatures rather than overfitting to fixed spatial layouts.

In addition to the visual results, we quantitatively analyzed VIN's performance on the full held-out set (20 WSIs). For evaluation, we operated at a region scale of $4096 \times 4096$ pixels rather than the $256 \times 256$ patches used for training. The rationale was to dampen cellular-level variability and emphasize tissue-level cues that are most relevant for margin localization in clinical practice. In routine use, the pathologist benefits from a WSI-level, low-resolution margin map, while the model internally leverages local, high-resolution patches with cellular and subcellular features to inform those global decisions; this multi-scale design explains the train–evaluate difference. Figure 4a summarizes the region-level results. We aggregated patch predictions within each $4096 \times 4096$ region using a voting scheme (see the Methods) to yield a single decision per region and visualized performance with a confusion matrix. Overall, the virtual inking model demonstrates an accuracy of ~73.3% for region-level classification across the test set. While patch-level analysis indicates a false-negative rate of ~36.7%, these errors manifest as intermittent signal discontinuities rather than structural omissions at the margin level. A voting- and stitching-based post-processing pipeline (see the Methods) aggregates patch decisions to reconstruct the linear trajectory of the cautery edge. Consequently, the final overlay presents a coherent, continuous boundary to the pathologist, suppressing isolated, noisy predictions that do not reflect the actual margin status. To support this claim, we conducted an additional manual evaluation on individual, continuous margins rather than localized small patches; see Supplementary Figure 2. In this evaluation, a board-certified pathologist (N.P.) manually labeled each consecutive margin based on the predictions of VIN for the 20 testing WSIs, which directly tested the clinical relevance of our method. We present the resulting confusion matrix in Fig. S2a in which each entry represents a single consecutive margin. We achieved an overall accuracy of ~74.3% and zero false positive boundaries. An example WSI is shown in Fig. S2b and S2c with 4 correct predictions and one false negative boundary. This analysis further demonstrates the potential clinical utility of VIN by providing accurate information at the margin level.

## Discussion

This work reported a virtual inking network that autonomously labels resection margins using a foundation model, thereby reducing reliance on inks and other physical labels. The workflow is free of hazardous reagents, preserves tissue, and demonstrates accurate, robust performance on human samples. In practice, VIN can be integrated into diagnostic review to generate a whole-slide margin map and to compute distances from putative malignant foci to the nearest cut surface, supporting margin-aware interpretation and case triage. The approach can also label resection margins in small, un-inked specimens that are subsequently found to harbor neoplastic tissue, providing a standardized virtual marker for documentation and measurement.

Another natural extension of this work is integration with virtual staining technologies[11–13], where the H&E-staining process of the tissue can also be conducted in a data-driven approach to virtually stain label-free images. Because cauterization induces immediate histomorphologic alterations, protein denaturation and cellular coagulation, with distinct autofluorescence signatures relative to remote, non-cauterized tissues[14], a multiplexed pipeline[15–17] is feasible: a single network model that ingests label-free microscopic images and simultaneously outputs virtually stained H&E and a virtual ink overlay at the cut surface. Such multiplexed inference would standardize both tissue contrast and edge delineation from the same acquisition, with the potential to streamline slide preparation and shorten time-to-result in settings where physical staining or inking is inconsistent. In addition, VIN might allow for the retrospective delineation of margins in small gastrointestinal polyps where carcinoma or high-grade dysplasia was incidentally identified. The current practice for these un-inked specimens often requires an extended resection in a separate procedure, as the tumor's distance from the margins cannot be reliably calculated, thus mandating the assumption that the tumor is close to the boundary. The extended resection reports can then be collected and reviewed to determine whether residual neoplastic tissue was present. Subsequently, the relationship between the distance from the margins on the original polyp (as calculated using VIN) and the presence of residual cancer in the extended resection can be assessed.

We selected HIPT[9] as the foundation feature extractor based on its computational efficiency and strong patch-level representation. As alternative foundation models continue to emerge[18–24] and multiple strategies exist for aggregating patch features into slide-level representations[9,25,26], VIN is positioned to benefit from these advances, specifically richer patch encoders and improved cross-patch aggregation. At present, inference proceeds in a sliding-window manner and does not explicitly impose slide-level priors—in other words, our inference does not make use of the fact that cautery follows a largely continuous trajectory along the cut edge and should not exhibit high spatial frequency oscillations. We anticipate that incorporating such constraints across adjacent patches, serial sections, or even three-dimensional tissue blocks[27,28] will further improve boundary continuity and robustness of our presented approach.

Alternative formulations other than patch-level classification are also possible. One approach is pixel-wise, high-resolution semantic segmentation[29,30]; another is line-detection at lower magnification[31,32]. We selected a patch-level classification design for two reasons. First, accurate recognition of cautery requires cellular- and subcellular-level cues that are best captured in high-resolution patches. Second, at the point of use, the pathologist benefits from a continuous, region-scale margin map rather than per-pixel decisions; cautery tends to form extended, contiguous edges, making patch-level outputs sufficient for reliable visualization and measurement while reducing sensitivity to spurious pixel-level noise.

As mentioned in our Results section, this study used experimentally induced *in situ* cauterization of tissue, which can yield shorter, fragmented cautery signatures and label variability across serial sections. These conditions may differ from the more continuous patterns typically produced during intraoperative resections. Future studies will validate VIN on routine surgical specimens across sites and tumor types, where we expect improved performance owing to longer, more contiguous cautery edges and reduced annotation ambiguity.

Overall, VIN provides a standardized, ink-free margin localization framework that can be translated onto routine digital pathology, with potential integration with virtual staining pipelines, incorporation of global spatial priors, and extension to multi-section or block-level analysis.

## Methods

## Sample preparation and data acquisition

Six human, non-neoplastic tonsils were subjected to experimentally induced *in situ* cauterization at the Department of Otolaryngology–Head & Neck Surgery, UC Davis Medical Center, under approval from the UC Davis Institutional Review Board (protocol #930499). Images of the six tonsil specimens are shown in Supplementary Figure S3. The specimens then underwent sectioning and embedding, yielding two tissue blocks per tonsil; thumbnails of these blocks are shown in Supplementary Figure S4. From each block, 10 serial sections were cut, totaling 120 tissue sections. Slides were stained with H&E and digitized to high-resolution brightfield WSIs using a scanning microscope (Axio Scan.Z1, Zeiss) with a 20×/0.8 NA Plan-Apochromat objective at the UCLA Translational Pathology Core Laboratory (TPCL). Manual annotation of cauterized areas was performed on the WSIs by a board-certified pathologist (N.P.) using the Automated Slide Analysis Platform (ASAP)[33]. Sparse coordinates were assigned to three categories: cautery (red), non-cautery (blue), and equivocal (green); equivocal regions were reserved for display only, and excluded from training and testing.

## Image preprocessing and label refinement

We collected 120 H&E-stained WSIs with coarse manual labels. To enable accurate and efficient annotation of histological features in WSIs and better align labels with the sample boundaries, we developed a semi-automated, interactive label-refinement tool using MATLAB (The MathWorks, Inc.; R2024) (Supplementary Figure S5). This methodology integrates manual guidance with automated edge detection to achieve precise boundary definitions. We start the interactive process by loading a 16x bilinear downsampled (compared to native resolution) image from the WSI structure. Then we correct the non-uniform illumination artifacts common in digital pathology and enhance local feature contrast using a background-subtraction method by dividing the source image by a Gaussian-blurred version of itself ($\sigma = 20$ pixels). To improve visual clarity, we applied a morphological closing operation with a 3-pixel-radius disk-shaped structuring element to fill minor holes and smooth the contours of tissue structures. After loading and preprocessing, we initiate an interactive loop where a human annotator provides initial guidance, which the tool then automatically refines. The process begins with the annotator drawing a freehand Region of Interest (ROI) around a specific histological feature using an assisted drawing tool. We create a binary mask to isolate the corresponding region in the contrast-enhanced image, followed by an edge-detection algorithm applied only within this masked region to identify high-gradient pixels that indicate the precise tissue boundary. Because the detected edge regions must be within the marked boundaries and form a continuous contour, a morphological closure operation is applied to the resulting edge map. Finally, the closed contour with specified continuity is used to determine the pixel coordinates.

Subsequently, the user assigns a morphological class to the annotation using a graphical user interface (Cautery, Non-Cautery, Equivocal). We repeat this process for all ROIs and store the refined label coordinates, which we later rasterize into images.

To focus the model on classifying whether an input patch represents cauterization or not, we filtered out background and intraparenchymal (non-edge) regions by design. First, each WSI was tiled into 4096 × 4096–pixel regions with 10% overlap. Next, the label coordinates were rasterized onto the image grid using the Bresenham algorithm[34], and any region containing no labeled pixels was discarded. For each remaining region, we further split it into 256 × 256–pixel patches. Each patch received a single class label based on pixel counts in the rasterized label mask. Patches were then categorized as cauterized or non-cauterized and compared with the ground truth. The ground-truth class for each patch was determined by a simple majority rule based on the rasterized pixel counts; whichever label contributed the greatest number of pixels defined the patch class.

## Network architecture, training and evaluation

Our virtual inking network contains a fixed feature extractor using a foundation model and a compact backend classifier to make the final predictions. The structure is illustrated in Figures 5a and 5 b. We used HIPT[9] as our feature extractor. HIPT contains multiple cascaded transformer architectures;[35,36] however, here we only use the frontend encoder, which encodes cellular-level features. The default input image shapes are 256 × 256 pixels, which can be altered by interpolating positional embeddings,[37] while we chose to use the native input resolution. We use only the classification token (CLS)[38] as the feature vector and discard all other local tokens, given the less demanding resolution requirements of our task.

To predict the final class of each input image patch, we kept the feature extractor fixed and optimized a compact classifier, which is a two-layer MLP. We kept the hidden dimension equal to the input dimension and used a ReLU[39] activation layer followed by a dropout layer[40] with 0.2 probability to mitigate overfitting. To avoid a feature-extraction bottleneck in training speed, we pre-extracted features and cached them for the shallow classifier training. The classifier was trained with an Adam optimizer[41] using cross-entropy loss with a learning rate of $5 \times 10^{-4}$ for up to 1,000 epochs; the checkpoint with the highest validation accuracy was selected as the final model. We used the numerically stable cross-entropy calculated as:

$$\text{CE}(y, z) = \max(z, 0) - zy + \log\left(1 + \exp(-|z|)\right)$$

where $z$ is the logit directly predicted by the network and $y$ is the label. For inference, we calculated the final class score using a sigmoid function:

$$P = \sigma(z) = \frac{1}{1 + \exp(-z)}$$

and assigned the final patch label using a threshold of 0.5, i.e., $\hat{y} = 1$ if $P \geq 0.5$ (cautery present) and $\hat{y} = 0$ otherwise, where $\hat{y}$ denotes the final predicted class.

For region-level evaluations (Fig. 4), we employed a voting-based scheme to generate a single prediction for each 4,096 × 4,096-pixel region, which was then used to populate the blind testing confusion matrices. We first performed inference at the patch level (256 × 256 pixels) by gridding each tissue region into 16 × 16 patches and obtaining a binary decision per patch (cauterization present/absent). The region label was determined by majority vote over the participating patches. Patches corresponding to inner/empty areas with no label did not participate in the voting, and all equivocal labels were excluded from evaluation. For visualizations in Figs. 2 and 3, each region is rendered uniformly according to the final voted decision of its parent region, and inner/empty areas are left blank. For the manual review in Supplementary Fig. S2, the pathologist was shown only low-magnification versions of each WSI with the VIN overlay; high-magnification views (zoom-ins) were disabled to avoid reliance on cellular-level cues. This design ensured that judgments reflected the VIN predictions at the margin scale, rather than independent re-annotation based on microscopic features.

Accuracy is calculated as:

$$\text{Accuarcy} = \frac{TP + TN}{TP + FP + TN + FN}$$

The false negative rate (FNR) is calculated as:

$$\text{FNR} = \frac{FN}{TP + FN}$$

Here $TP, TN, FP,$ and $FN$ denote the numbers of true positives, true negatives, false positives, and false negatives, respectively (with "cauterized" as the positive class).

## Acknowledgments

The authors acknowledge the support of the NIH P41, the National Center for Interventional Biophotonic Technologies (P41EB032840). N.P. is partially supported by a PhRMA Foundation Translational Medicine Postdoctoral Fellowship.

## References

1. Shah, A. K. Postoperative pathologic assessment of surgical margins in oral cancer: A contemporary review. *J. Oral Maxillofac. Pathol.* **22**, 78–85 (2018).

2. Houssami, N., Macaskill, P., Luke Marinovich, M. & Morrow, M. The Association of Surgical Margins and Local Recurrence in Women with Early-Stage Invasive Breast Cancer Treated with Breast-Conserving Therapy: A Meta-Analysis. *Ann. Surg. Oncol.* **21**, 717–730 (2014).

3. Rosai, J. *Rosai and Ackerman's Surgical Pathology e-Book*. (Elsevier Health Sciences, 2011).

4. Sutton, D. N., Brown, J. S., Rogers, S. N., Vaughan, E. D. & Woolgar, J. A. The prognostic implications of the surgical margin in oral squamous cell carcinoma. *Int. J. Oral Maxillofac. Surg.* **32**, 30–34 (2003).

5. Sunkara, P. R., Graff, J. T. & Cramer, J. D. Association of surgical margin distance with survival in patients with resected head and neck squamous cell carcinoma: a secondary analysis of a randomized clinical trial. *JAMA Otolaryngol. Neck Surg.* **149**, 317–326 (2023).

6. Seethala, R. R. *et al.* Protocol for the Examination of Specimens from Patients with Cancers of the Oral Cavity. (2021).

7. Williams, A. S. & Dakin Haché, K. Variable fidelity of tissue-marking dyes in surgical pathology. *Histopathology* **64**, 896–900 (2014).

8. Taqi, S. A., Sami, S. A., Sami, L. B. & Zaki, S. A. A review of artifacts in histopathology. *J. Oral Maxillofac. Pathol. JOMFP* **22**, 279 (2018).

9. Chen, R. J. *et al.* Scaling vision transformers to gigapixel images via hierarchical self-supervised learning. in *Proceedings of the IEEE/CVF Conference on Computer Vision and Pattern Recognition* 16144–16155 (2022).

10. Rumelhart, D. E., Hinton, G. E. & Williams, R. J. *Learning Internal Representations by Error Propagation*. https://apps.dtic.mil/sti/html/tr/ADA164453/ (1985).

11. Bai, B. *et al.* Deep learning-enabled virtual histological staining of biological samples. *Light Sci. Appl.* **12**, 57 (2023).

12. Rivenson, Y. *et al.* Virtual histological staining of unlabelled tissue-autofluorescence images via deep learning. *Nat. Biomed. Eng.* **3**, 466–477 (2019).

13. Rivenson, Y., de Haan, K., Wallace, W. D. & Ozcan, A. Emerging Advances to Transform Histopathology Using Virtual Staining. *BME Front.* **2020**, 9647163 (2020).

14. Lagarto, J. L. *et al.* Electrocautery effects on fluorescence lifetime measurements: An in vivo study in the oral cavity. *J. Photochem. Photobiol. B* **185**, 90–99 (2018).

15. Zhang, Y. *et al.* Digital synthesis of histological stains using micro-structured and multiplexed virtual staining of label-free tissue. *Light Sci. Appl.* **9**, 78 (2020).

16. Zhang, Y. *et al.* Deep learning-enabled virtual multiplexed immunostaining of label-free tissue for vascular invasion assessment. Preprint at https://doi.org/10.48550/arXiv.2508.16209 (2025).

17. Yang, X. *et al.* Virtual birefringence imaging and histological staining of amyloid deposits in label-free tissue using autofluorescence microscopy and deep learning. *Nat. Commun.* **15**, 7978 (2024).

18. Chen, R. J. *et al.* Towards a general-purpose foundation model for computational pathology. *Nat. Med.* **30**, 850–862 (2024).

19. Zimmermann, E. *et al.* Virchow2: Scaling Self-Supervised Mixed Magnification Models in Pathology. Preprint at http://arxiv.org/abs/2408.00738 (2024).

20. Xu, H. *et al.* A whole-slide foundation model for digital pathology from real-world data. *Nature* **630**, 181–188 (2024).

21. bioptimus. releases/models/h-optimus/v0 at main · bioptimus/releases. *GitHub* https://github.com/bioptimus/releases/tree/main/models/h-optimus/v0.

22. Lu, M. Y. *et al.* A visual-language foundation model for computational pathology. *Nat. Med.* **30**, 863–874 (2024).

23. ai, kaiko *et al.* Towards Large-Scale Training of Pathology Foundation Models. Preprint at https://doi.org/10.48550/arXiv.2404.15217 (2024).

24. Ding, T. *et al.* A multimodal whole-slide foundation model for pathology. *Nat. Med.* 1–13 (2025) doi:10.1038/s41591-025-03982-3.

25. Ilse, M., Tomczak, J. & Welling, M. Attention-based deep multiple instance learning. in *International conference on machine learning* 2127–2136 (PMLR, 2018).

26. Ding, J. *et al.* LongNet: Scaling Transformers to 1,000,000,000 Tokens. Preprint at https://doi.org/10.48550/arXiv.2307.02486 (2023).

27. Chen, C.-H. *et al.* Shrinkage of head and neck cancer specimens after formalin fixation. *J. Chin. Med. Assoc.* **75**, 109–113 (2012).

28. Gao, G. *et al.* Abstract 2437: Deep-learning triage of 3D pathology datasets for accurate and efficient pathologist assessments. *Cancer Res.* **85**, 2437 (2025).

29. Deng, R. *et al.* Segment anything model (sam) for digital pathology: Assess zero-shot segmentation on whole slide imaging. in *IS&T International Symposium on Electronic Imaging* vol. 37 COIMG-132 (2025).

30. Cheng, B., Misra, I., Schwing, A. G., Kirillov, A. & Girdhar, R. Masked-Attention Mask Transformer for Universal Image Segmentation. in 1290–1299 (2022).

31. Carion, N. *et al.* End-to-End Object Detection with Transformers. in *Computer Vision – ECCV 2020* (eds Vedaldi, A., Bischof, H., Brox, T. & Frahm, J.-M.) 213–229 (Springer International Publishing, Cham, 2020). doi:10.1007/978-3-030-58452-8_13.

32. Xu, Y., Xu, W., Cheung, D. & Tu, Z. Line segment detection using transformers without edges. in *Proceedings of the IEEE/CVF Conference on Computer Vision and Pattern Recognition* 4257–4266 (2021).

33. ASAP - Automated Slide Analysis Platform. https://computationalpathologygroup.github.io/ASAP/.

34. Bresenham, J. E. Algorithm for computer control of a digital plotter. in *Seminal graphics* 1–6 (ACM, New York, NY, USA, 1998). doi:10.1145/280811.280913.

35. Vaswani, A. *et al.* Attention is all you need. *Adv. Neural Inf. Process. Syst.* **30**, (2017).

36. Dosovitskiy, A. An image is worth 16x16 words: Transformers for image recognition at scale. *ArXiv Prepr. ArXiv201011929* https://files.ryancopley.com/Papers/2010.11929v2.pdf (2020).

37. Touvron, H. *et al.* Training data-efficient image transformers & distillation through attention. in *International conference on machine learning* 10347–10357 (PMLR, 2021).

38. Devlin, J., Chang, M.-W., Lee, K. & Toutanova, K. Bert: Pre-training of deep bidirectional transformers for language understanding. in *Proceedings of the 2019 conference of the North American chapter of the association for computational linguistics: human language technologies, volume 1 (long and short papers)* 4171–4186 (2019).

39. Nair, V. & Hinton, G. E. Rectified linear units improve restricted boltzmann machines. in *Proceedings of the 27th International Conference on International Conference on Machine Learning* 807–814 (Omnipress, Madison, WI, USA, 2010).

40. Srivastava, N., Hinton, G., Krizhevsky, A., Sutskever, I. & Salakhutdinov, R. Dropout: a simple way to prevent neural networks from overfitting. *J Mach Learn Res* **15**, 1929–1958 (2014).

41. Kingma, D. P. & Ba, J. Adam: A Method for Stochastic Optimization. *ArXiv14126980 Cs* http://arxiv.org/abs/1412.6980 (2017).

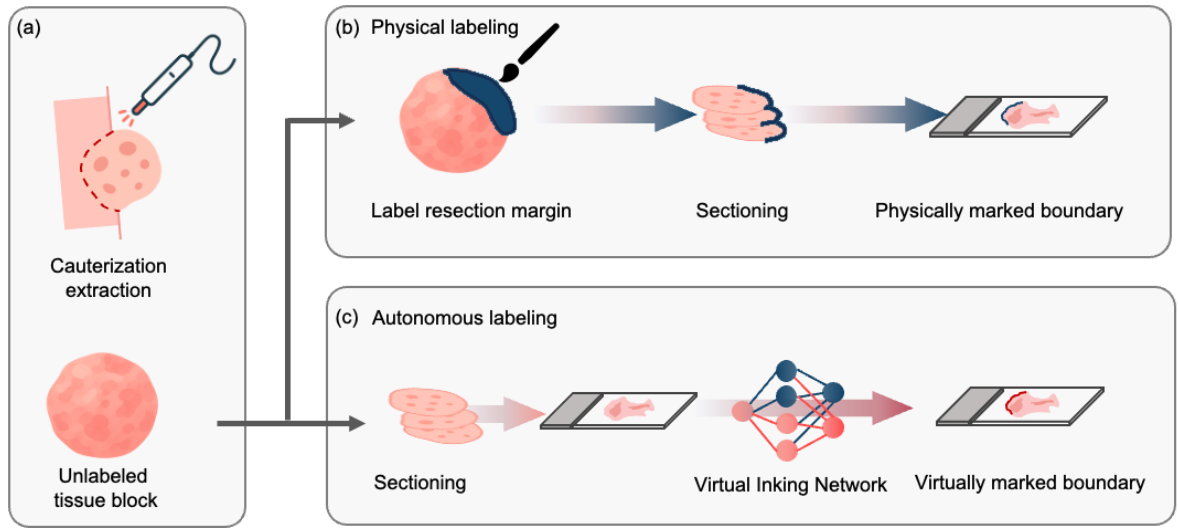

**Figure 1. Autonomous labeling of surgical resection margins.** (a) Tissue extraction with electrocautery. (b) Traditional workflow where ink is physically applied to mark resection margins. (c) Autonomous labeling using the virtual inking network (VIN).

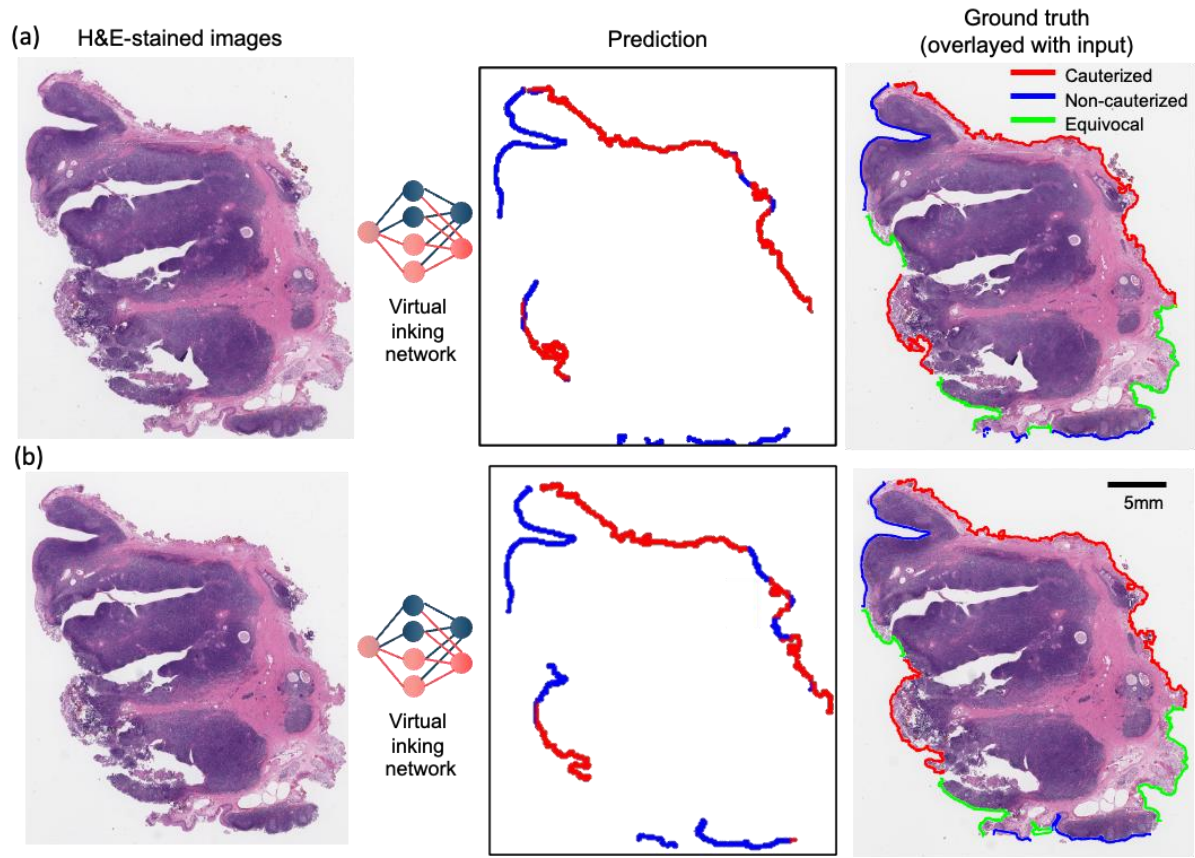

**Figure 2. Whole-slide level evaluation of virtual inking network.** We blindly tested two whole-slide images (a,b) and show the input H&E images on the left panel with the virtual inking network predictions in the middle and the ground truth labels overlaid on the H&E images on the right.

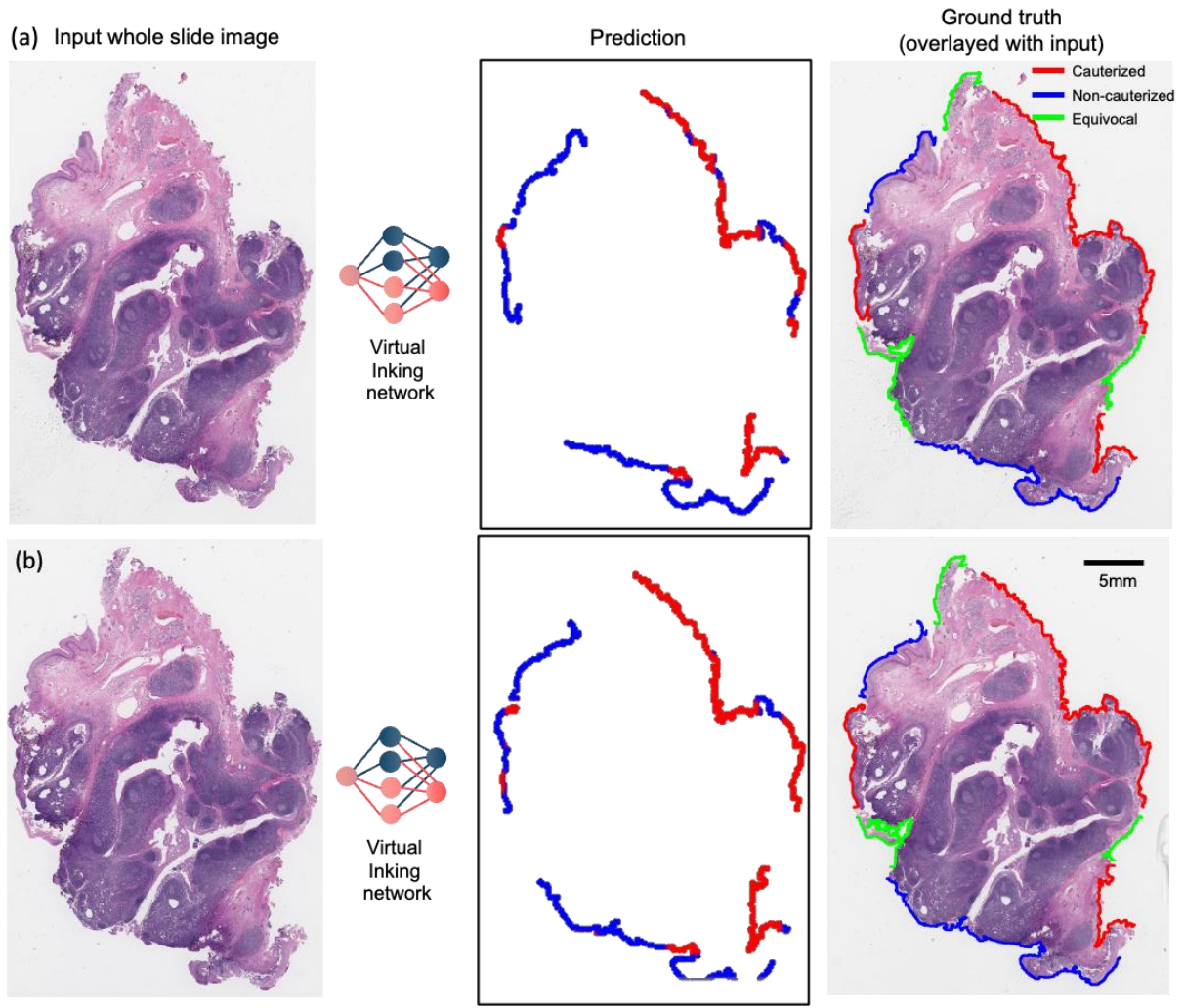

**Figure 3. Whole-slide level evaluation of virtual inking network on another tissue block.** We blindly tested two whole-slide images (a,b) and show the input H&E images on the left panel with the virtual inking network predictions in the middle and the ground truth labels overlaid on the H&E images on the right.

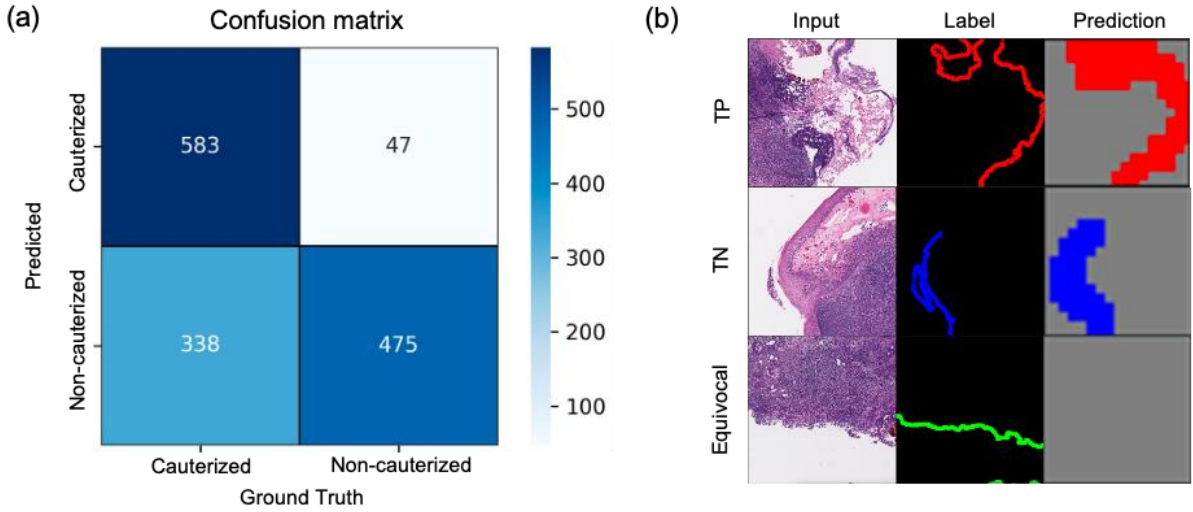

**Figure 4. Quantitative evaluation of virtual inking network performance.** We show the confusion matrix in (a) and show some prediction examples in (b). TP: true positive, TN: true negative.

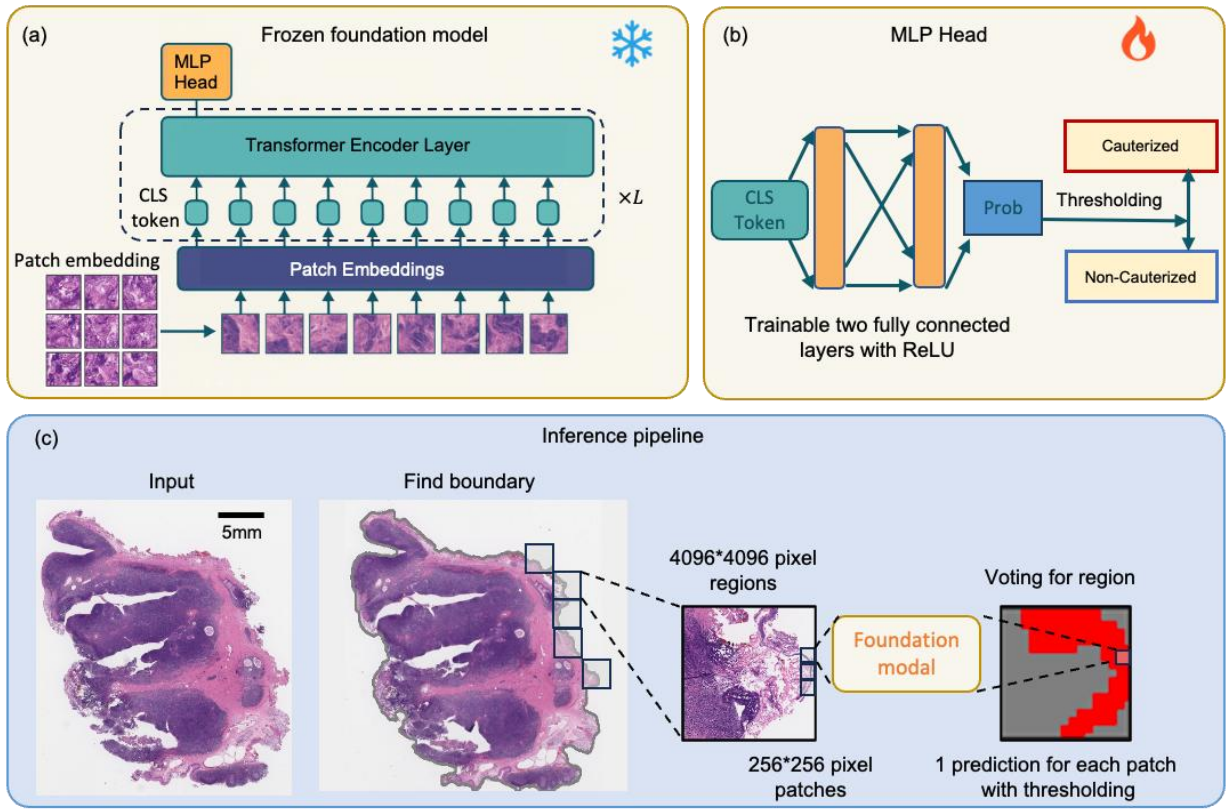

**Figure 5. Virtual inking network architecture.** (a) Frozen foundation model as a feature extractor. (b) Trainable shallow classification head. (c) The inference pipeline of the virtual inking network.